\documentclass[letterpaper, 10 pt, conference]{ieeeconf}  % Comment this line out if you need a4paper

\IEEEoverridecommandlockouts                              % This command is only needed if 
\usepackage{xcolor}

\usepackage{graphicx}

\usepackage{gensymb}
\usepackage[hyphens]{url}
\usepackage{multirow}
\usepackage{multicol}
\usepackage{tabularx}
\usepackage{dsfont}
\usepackage{url}
\usepackage{color}
\usepackage{subcaption}
\usepackage{caption}
\usepackage{booktabs}
\usepackage{dcolumn}
\usepackage{amssymb}
\usepackage{hyperref}
\usepackage{amsmath}
\usepackage{colortbl}
\usepackage{tikz}
\colorlet{lightgray}{gray!20}

\hypersetup{
    colorlinks=false,
    linkcolor=blue,
    filecolor=magenta,      
    urlcolor=cyan,
    pdftitle={Overleaf Example},
    pdfpagemode=FullScreen,
    }

\title{\LARGE \bf
Bridging the Domain Gap for Multi-Agent Perception }

\author{Runsheng Xu$^{1}$, Jinlong Li$^{2}$, Xiaoyu Dong$^{3}$, Hongkai Yu$^{2}$, Jiaqi Ma$^{1*}$
\thanks{$^{1}$University of California, Los Angeles, UCLA Mobility Lab. {\{\tt\small{rxx3386, jiaqima}\}}{\tt\small{@ucla.edu}}}
\thanks{$^{2}$Cleveland State University,  Cleveland Vision \& AI Lab. {\tt\small{j.li56@vikes.csuohio.edu, h.yu19@csuohio.edu}}}
\thanks{$^{3}$Northwestern University. }
\thanks{*Corresponding Author}}
\begin{document}

 \maketitle
\thispagestyle{empty}
\pagestyle{empty}

%%%%%%%%%%%%%%%%%%%%%%%%%%%%%%%%%%%%%%%%%%%%%%%%%%%%%%%%%%%%%%%%%%%%%%%%%%%%%%%%
\begin{abstract}
Existing multi-agent perception algorithms usually select to share deep neural features extracted from raw sensing data between agents, achieving a trade-off between accuracy and communication bandwidth limit. However, these methods assume all agents have identical neural networks, which might not be practical in the real world. The transmitted features can have a large domain gap when the models differ, leading to  a dramatic performance drop in multi-agent perception. In this paper, we propose the first lightweight framework to bridge such domain gaps for multi-agent perception, which can be a plug-in module for most of the existing systems while maintaining confidentiality. Our framework consists of a learnable feature resizer to align features in multiple dimensions and a sparse cross-domain transformer for domain adaption. Extensive experiments on the public multi-agent perception dataset V2XSet have demonstrated that our method can effectively bridge the gap for features from different domains and outperform other baseline methods significantly by at least 8\% for point-cloud-based 3D object detection.

% Recent literature has demonstrated that by sharing the visual features from nearby autonomous vehicles and infrastructure, perception performance can reach a new level. 

% dataset contribution
% proposed new attentive intermediate fusion model 
% development kit and data are available online 

% experiment part -- need data analysis

\end{abstract}

%%%%%%%%%%%%%%%%%%%%%%%%%%%%%%%%%%%%%%%%%%%%%%%%%%%%%%%%%%%%%%%%%%%%%%%%%%%%%%%%
\section{INTRODUCTION}
\begin{figure}[!t]
\centering
\subfloat[Multi-agent perception pipeline in the context of V2X perception]{%
  \includegraphics[height=0.45\columnwidth,width=1\columnwidth]{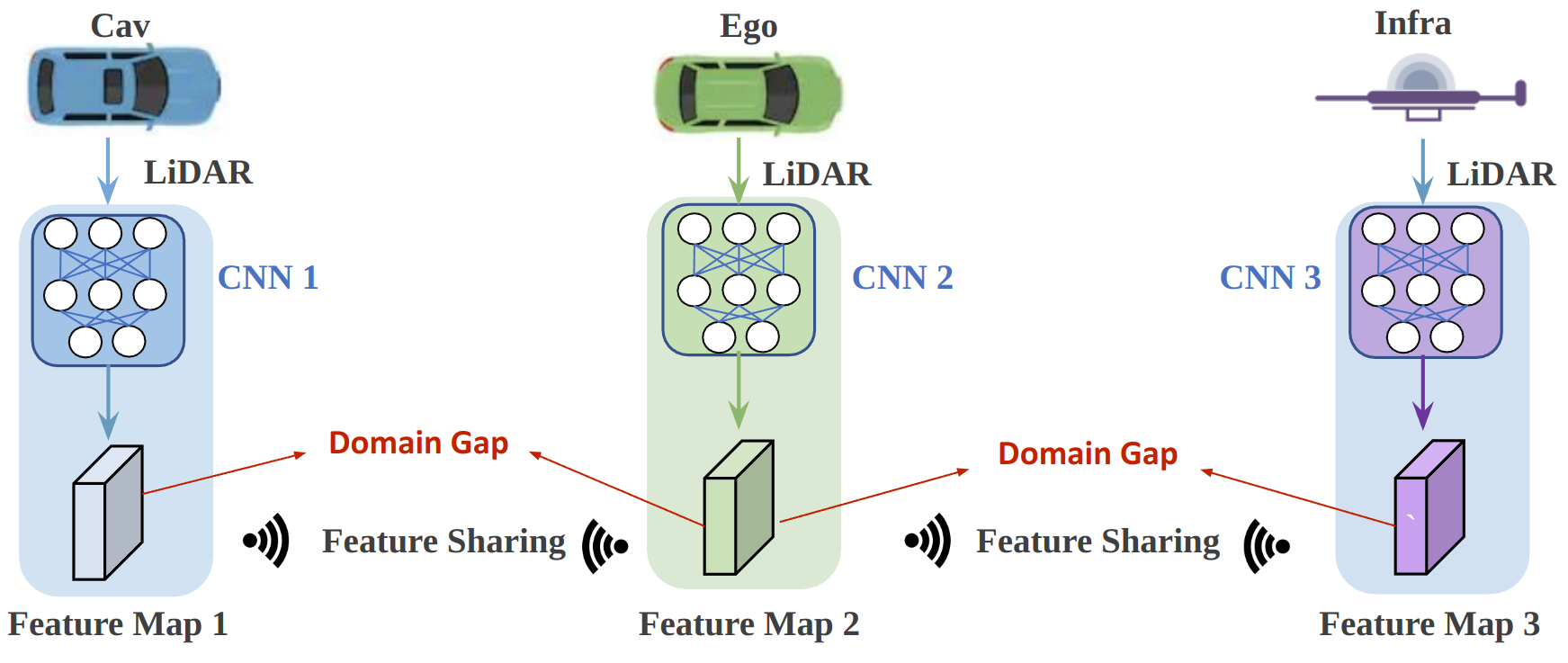}%
}
\hfil
\subfloat[PointPillar feature map]{%
  \includegraphics[height=0.25\columnwidth,width=0.5\columnwidth]{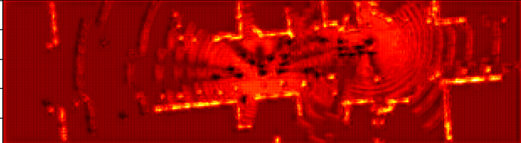}%
}
\hfil
\subfloat[VoxelNet feature map]{%
  \includegraphics[height=0.25\columnwidth,width=0.5\columnwidth]{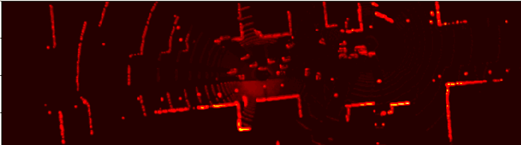}%
}
\caption{\textbf{Illustration of domain gap of different feature maps for multi-agent perception.}. Here we use  V2X cooperative perception in autonomous driving as an example. (a) Ego vehicle receives the shared feature maps from other CAV and infrastructure with different CNN models, which causes domain gaps. (b) Visualization of feature map from ego, which is extracted from PointPillar~\cite{lang2019pointpillars}. (c) Feature map from CAV, which is extracted from VoxelNet~\cite{zhou2018voxelnet}. Brighter pixels represent higher feature values.}
\label{fig:comparison}
\vspace{-4mm}
\end{figure}

Recent studies have demonstrated that by leveraging Vehicle-to-Everything~(V2X) communication technology to share visual information, the multi-agent perception system can significantly improve the performance of the single-agent system by seeing through occlusions and perceiving longer range~\cite{xu2022opv2v, xu2022cobevt, xu2022v2xvit, hua2019hierarchical, li2021learning, chen2023dynamic, yuan2022keypoints, lei2022latency}. Instead of sharing raw sensing data or detected outputs, state-of-the-art methods usually share the intermediate neural features computed from the sensor data,  as they can achieve the best trade-off between accuracy and bandwidth requirements~\cite{wang2020v2vnet, xu2022opv2v}. Furthermore, transmitted intermediate features are more robust to the GPS noise and communication delay~\cite{xu2022v2xvit, liu2021automated, lei2022latency}. Despite the advancements in intermediate fusion strategy,  previous methods conduct experiments under a strong assumption that all agents are equipped with identical neural networks to extract neural features. This overlooks a critical fact: deploying the same model for all agents is unrealistic, especially for connected autonomous driving~\cite{song2022federated, song2022edge}. For example, as shown in  (a) from Fig.~\ref{fig:comparison}, the detection models on connected automated vehicles~(CAV) and infrastructure products of distinct companies are usually dissimilar. Even for the same company, diverse detection models may exist due to the different on-vehicle software versions. When the shared features come from different backbones, a noticeable domain gap exists, which can easily diminish the benefits of collaborations.

In this paper,  we dive into this unsolved and practical problem in multi-agent perception, especially for autonomous driving. We first carefully investigate the domain gap of different feature maps and then propose our framework based on the analysis. Fig.~\ref{fig:comparison} shows intermediate feature representations obtained from two distinct point cloud based 3D object detection backbones, PointPillar~\cite{lang2019pointpillars} and VoxelNet~\cite{zhou2018voxelnet}, in the same scenario. We apply the same techniques as \cite{zagoruyko2016paying} to make the visualization informative by summing up all channels' absolute value together.  In general, we can observe the features are dissimilar in three aspects: 
\begin{itemize}
    \item \textbf{Spatial resolution.} Because of the different voxelization parameters, LiDAR cropping range,  and downsampling layers, the spatial resolutions are different.
    \item \textbf{Channel number.} The channel dimensions are distinct due to the difference in convolution layers' settings.
    \item \textbf{Patterns.} As Fig.~\ref{fig:comparison} shows, PointPillar and VoxelNet have the opposite patterns: The object positions have relatively low values on the feature map for PointPillar but high values for VoxelNet. 
    
\end{itemize}

\begin{figure*}[!t]
\centering
\includegraphics[width=0.83\linewidth]{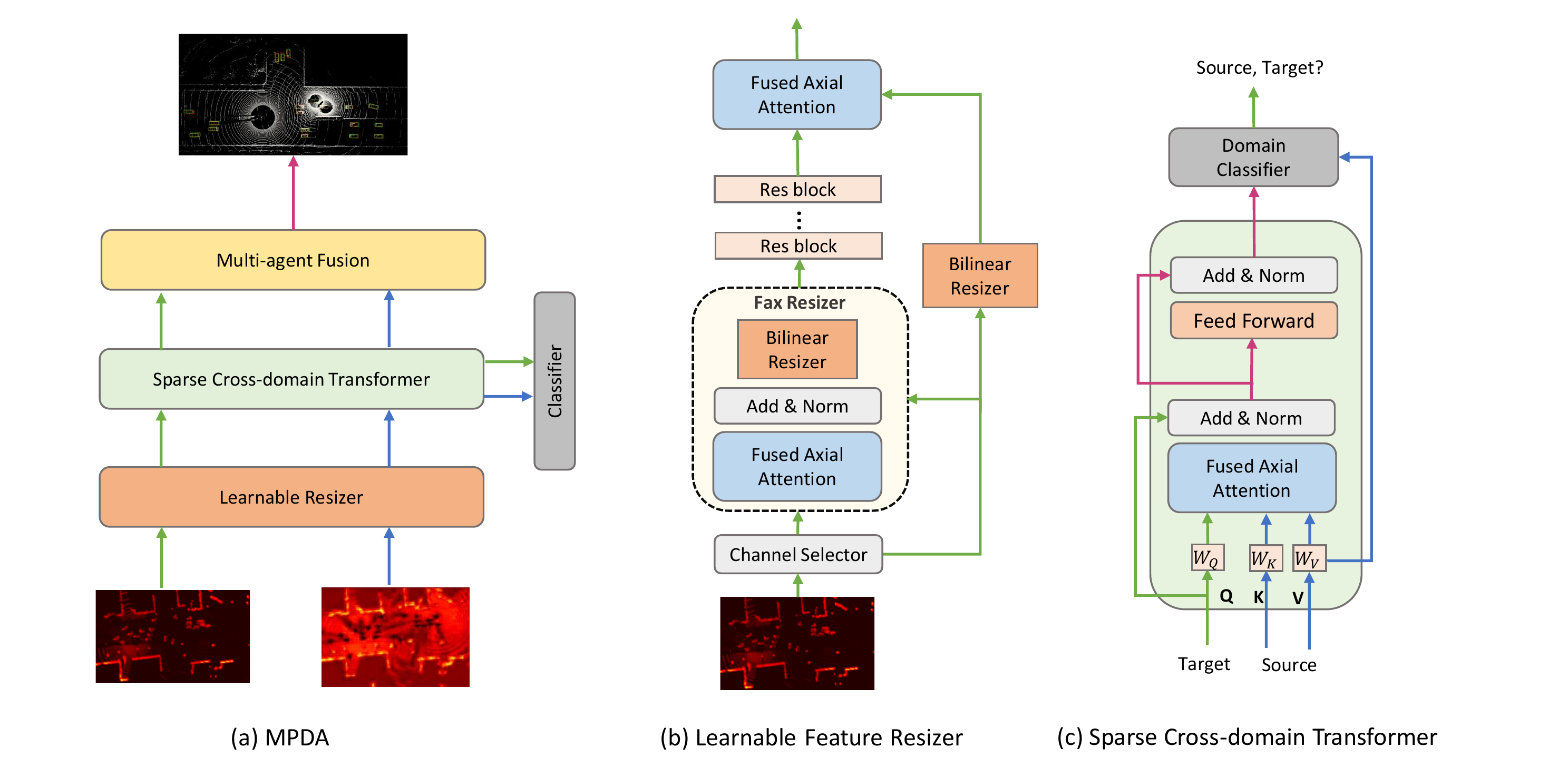}
\caption{\textbf{The overview  and core components of our framework.} Our MPDA first aligns feature dimensions through a learnable feature resizer and then unifies the pattern through the sparse cross-domain transformer.}
\label{fig:arch}
\vspace{-3mm}
\end{figure*}

To address the three dominant distinctions, we present the first \textbf{M}ulti-agent \textbf{P}erception \textbf{D}omain \textbf{A}daption framework, dubbed as MPDA, to bridge the domain gap. Fig.~\ref{fig:arch} depicts the overall architecture. Specifically,  two components, namely Learnable Resizer and Sparse Cross-Domain Transformer, are proposed. As multiple factors could cause different spatial resolutions, we argue that using rudimentary resizing algorithms such as bilinear and nearest interpolation may cause severe misalignment. Therefore, we propose to resize the received intermediate features in a learnable way and optimize with the multi-agent fusion algorithms jointly to improve the detection performance.
Moreover, aligning the channel dimension by simply dropping channels can potentially lead to losing important information; thus, our resizer also includes a learnable channel selector to alleviate such loss. To diminish the pattern disparity, the sparse cross-domain transformer will efficiently reason the received and ego features locally and globally and generates domain-invariant representations by adversarially fooling a domain classifier. Finally, the state-of-the-art multi-agent fusion algorithm V2X-ViT~\cite{xu2022v2xvit} is utilized to fuse information across multiple agents. Since the framework does not require any key information from other models~(e.g., model type, parameters), it can maintain confidentiality. We conduct extensive experiments on the public dataset V2XSet~\cite{xu2022v2xvit}, and the result demonstrates that our framework can increase the accuracy of V2X-ViT by at least 8\% under various realistic settings. Overall, our contributions are summarized as follows:

\begin{itemize}
    \item We pioneer the domain gap identification~(spatial resolution, channel number, pattern) in multi-agent perception and propose a new Multi-agent Perception Domain Adaption~(MPDA) framework, which is the \textbf{first work to bridge the domain gap for multi-agent perception}.  
    \item We present a novel Learnable Resizer to better align spatial and channel features from other agents in an adaptive way. 
    \item We propose a sparse cross-domain transformer that can efficiently unify the feature patterns from various agents. 
    \item The proposed MPDA framework can be easily combined with other multi-agent fusion algorithms and does not require confidential model information from other agents. Extensive experiments on the public dataset V2XSet demonstrate that our method achieves the best performance with real-time performance.
\end{itemize}

\section{Related Work}
\noindent\textbf{Multi-Agent Perception: } Despite the great progress in autonomous driving in the past years, there still exist many challenges that single-vehicle systems can not get over~\cite{9352029, 9684293, 9851671,9495943,valiente2022robustness,calvert2020conceptual,shet2021cooperative,ferrara2022multi,khalil2022licanet,kitajima2022nationwide}. Especially, the single-agent perception systems suffer severely from occlusions and limitations in sensor range \cite{hu2022investigating}. Multi-agent perception was born to alleviate such pains. As a pioneering work, V2VNet~\cite{wang2020v2vnet} first proposes the intermediate fusion approach, in which all agents should broadcast the features extracted from the raw point cloud to achieve the trade-off between bandwidth requirement and accuracy. Following this design ethos, OVP2V~\cite{xu2022opv2v} uses a single-head self-attention module to fuse the received intermediate features. DiscoNet~\cite{li2021learning} utilizes a graph neural network and knowledge distillation to aggregate the shared representations. Recently, V2X-ViT~\cite{xu2022v2xvit} first proposed to use a vision transformer for multi-agent perception and achieves robust performance under GPS error and communication delay. \cite{su2022uncertainty} captures both aleatoric and epistemic uncertainties with one inference pass and tailors a moving block bootstrap algorithm with direct modeling of the multivariant Gaussian distribution of each corner of the bounding box. The method can be used with different collaborative object detectors and helps to improve safety-critical systems such as CAVs. Despite the prominent performance achieved by these methods, none of them consider the realistic domain gap issue caused by the model discrepancy. We aim to fill such a gap in this paper.

% early fusion: 

% late fusion: CAR2X-based perception (temporal spatial alignment)

% intermediate fusion: 

\noindent\textbf{Transformers in Vision: } Since Dosovitskiy \textit{et~al.}~\cite{dosovitskiy2020image} successfully adapt the Transformer architecture~\cite{vaswani2017attention} into the computer vision area by regarding image patches as visual words, Vision Transformers~(ViT) have gained increasing attention~\cite{ tracking, end2end, proto}. For example,  \cite{liu2022yolov5} shows that by applying a 3D attention mechanism, the object detection performance can outperform traditional convolutional neural networks.  Despite obtaining great benefits from global interactions, the full attention in ViT~\cite{vaswani2017attention, dosovitskiy2020image} usually requires large computation resources. To avoid such costs, recent methods~\cite{liu2021swin,fan2022svt,tu2022maxvit} have explored different sparse attention mechanisms such as local and sparsely global schemes. In this work, we adapt an efficient 3D attention called Fused Axial Attention~(FAX) in CoBEVT~\cite{xu2022cobevt} to our domain adaption framework as it already shows excellent efficiency on multi-agent fusion.

\noindent\textbf{Domain Adaptation: }
Due to the time consumption of data annotation and the domain gap between different domains, domain adaptation is utilized to solve these problems by adapting the model trained on a labeled source domain to address an unlabeled target domain. Recent works on domain adaptation mainly address different computer vision tasks~\cite{song2020multi,shao2021fadacs,du2021cross,li2021domain,fu2021let,xu2021cdtrans,yao2021multi,fan2022reconstruction,fan2021deep,yang2020dpgn}.
% , such as image classification~\cite{du2021cross,xu2021cdtrans,you2019universal,xiao2021dynamic} and object detection~\cite{hsu2020progressive,li2021domain,fu2021let,gu2021pit,yao2021multi}. 
In domain adaptation, to minimize the domain shift between different domains,  feature distribution can be aligned in common levels: domain level~\cite{long2015learning, sun2016deep,yu2017dilated} and category level~\cite{xu2021cdtrans,du2021cross,zhang2019domain, hu2020dasgil,Yang2022Unsupervised}. 
Domain level alignment generally involves minimizing some measure of distance between the source and target feature distributions like maximum mean discrepancy~\cite{long2015learning}. \cite{9998556} proposes a novel prototype-based shared-dummy classifier (PSDC) model to address the challenges of open-set domain adaptation, including distinguishing between unknown target instances and shared classes and aligning shared class prototypes, which outperforms existing methods on several datasets. \cite{sanqing2022BMD} proposes a novel class-Balanced Multicentric Dynamic prototype (BMD) strategy for the Source-free Domain Adaptation (SFDA) task to adapt pre-trained source models to the target domain without accessing the well-labeled source data. The proposed BMD strategy avoids the gradual dominance of easy-transfer classes on prototype generation, introduces a novel inter-class balanced sampling strategy, and incorporates dynamic network update information during model adaptation.
While category level alignment aligned each category distribution between source and target domain using an adversarial manner between the feature extractor and domain classifiers.
A fine-grained alignment leads to more accurate distribution alignment in the same label space. 
% Du et al.~\cite{du2021cross} propose a cross-domain gradient discrepancy minimization (CGDM) method to minimize the gradient discrepancy between two domains to achieve a better distribution alignment at category level. 
Xu et al.~\cite{xu2021cdtrans} adopted Transformers for category-level domain adaptation to show great potential in image classification. In this paper, similar to~\cite{xu2021cdtrans},  a sparse cross-domain Transformer is proposed to unify the feature patterns from different agents.

\noindent\textbf{Learnable Resizer:} 
\cite{talebi2021learning} first comes up the concept of learnable resizer for image classifications. Instead of using rudimentary interpolation, they employ a convolution neural network to resize the RGB images for classification and jointly train with vision models. Our learnable feature resizer is inspired by this work but differs in three major aspects: 1) We investigate an unexplored practical application scenario for a learnable resizer -- domain adaption for multi-agent perception. 2) Our resizing target is the LiDAR feature, which is more sparse than images. Therefore, instead of using a pure convolution neural network, we integrate our resizer with a sparse transformer. 3) Besides resizing the spatial dimension, we also embed a simple but effective algorithm to resize the channel dimension to the required number.

\section{Methodology}
In this paper, we consider a realistic scenario for multi-agent perception, where each agent in the collaboration may be equipped with a separate model and transmit visual features with domain discrepancy. We mainly focus on the cooperative perception task of LiDAR-based 3D object detection for autonomous driving, where the agents are connected to autonomous vehicles and intelligent roadside infrastructure, but our framework is generally-applicable to other multi-agent perception applications as long as they broadcast neural features for collaborations. Since we focus on the problem of domain gaps in this work, we assume the relative poses between agents are accurate and no communication delay exists.

Fig.~\ref{fig:arch}(a) shows the overall architecture of our MPDA, which consists of 1)  a learnable feature resizer, 2) a sparse cross-domain transformer, 3) a domain classifier, and 4) multi-agent feature fusion. In this section, we will describe the details of each module.

\subsection{Learnable Feature Resizer}
We regard the feature maps  computed locally on ego vehicle as source domain features $F_S \in \mathbb{R}^{1 \times H_S \times W_S \times C_S}$ and received features from other agents as target domain features $F_T \in \mathbb{R}^{N \times H_T \times W_T \times C_T}$, where $N$ is the number of other collaborators/agents, $H$ is the height, $W$ is the width, $C$ is the channel number, and $H_S \neq H_T, W_S \neq W_T, C_S \neq C_T$.  The goal of our feature resizer $\Phi$ is to align the dimensions of the source domain feature with the target domain in a learnable way:
\begin{equation}
    F_T^{'} = \Phi(F_T), \; \text{s.t.} \; F_T^{'} \in \mathbb{R}^{N \times H_S \times W_S \times C_S}.
\end{equation}
We jointly train $\Phi$ with multi-agent detection models so it can intelligently learn the optimal approach to resize the features, which is fundamentally different from the naive resizing method such as bilinear interpolation. The architecture of our learnable feature resizer is designed as Fig~\ref{fig:arch}(b) shows, which includes four major components: channel aligner, FAX resizer, skip connection, and res-block. 

\noindent\textbf{Channel Aligner:} We use a simple $1\times 1$ convolution layer to align the channel dimension, whose input channel number is $C_{in} = 2C_S$ and outputs $C_S$ channels. When $C_T > C_{in}$, we randomly drop $C_{in}-C_T$ channels and apply the $1 \times 1$ convolution layer to obtain a new feature. We repeat this process on $F_T$ for $n$ times to get features with ${n \times H_T \times W_T \times C_S}$ dimensions and average them along the first dimension. In this way, we ameliorate the loss of information due to channel dropping.  When $C_T < C_{in}$, we perform padding with randomly selected channels from $F_T$ to meet the required input channel number for the $1 \times 1$ convolution.

\noindent\textbf{FAX  Resizer:} To search for the optimal resizing solution, the neural network is supposed to have a large receptive field to gain the global information and pay attention to details to capture the critical object information. Since LiDAR features are usually sparse due to empty voxels, applying large-kernel convolution to get global information may diffuse the meaningless information to the important area. Therefore,  we apply the fused axial~(FAX) attention block~\cite{xu2022cobevt} before bilinear resizing to fetch better feature representations. FAX sparsely employs  local window and grid attention to efficiently capture global and local interactions. More importantly, it can discard empty voxels through a dynamic attention mechanism to eliminate their potential negative effects. After FAX, a bilinear resizer is implemented to reshape the feature map to the same spatial dimension as the source feature map. Compared to simple bilinear interpolation, our FAX resizer can adjust the input features first to avoid misalignment and distortion issues during resizing.

\noindent\textbf{Skip connection:} We also employ the bilinear feature resizing method in the skip connection to make learning easier.

% After channel alignment, we firstly apply the fused axial attention block~\cite{xu2022cobevt} to 

% capture the global interactions and  
% Before resizing the feature bilinearly, we first apply the fused axial attention block~\cite{xu2022cobevt} to capture the global

\noindent\textbf{Res-Block:} We implement standard residual blocks~\cite{he2016deep}  $r$ times after resizing the feature maps to further refine them. 

% Note that for simplicity, we assume there is only one target domain, but our learnble feature resizer can also handle multiple target domains as it accepts any feature size. 

\subsection{Sparse Cross-Domain Transformer}
After retrieving the resized feature $F_T^{'}$, we need to convert its pattern to be indistinguishable from the domain classifier to obtain the domain-invariant  features. To reach this goal, we need to effectively reason the correlations between $F_T^{'}$ and $F_S$ both locally and globally. Therefore, we propose the sparse cross-domain transformer, which enjoys the benefits of dynamic and global attention brought by the transformer architecture while avoiding expensive computation. Fig.~\ref{fig:arch}(c) shows the details of our proposed architecture. We first apply different convolution layers $W_Q, W_K, W_V$ on $F_T^{'}$ and $F_S$ to obtain query, key, and value, respectively. Then  the query from the target domain and key/value from the source domain will be fed into the FAX block, capturing sparsely local and global spatial interactions across target and source domain features. Finally, a standard feed-forward neural network~(FFN) is implemented to refine the interacted feature further. The whole process can be formulated as below:
\begin{align}
    Q = W_Q(F_T^{'}), \quad K = W_K(F_S), \quad V = W_V(F_S),  \\
    \hat{F_T^{'}} = Q+LN(FAX(Q, K, V)), \\
    F_T^{''} = \hat{F_T^{'}}+LN(FFN(\hat{F_T^{'}})), 
\end{align}
where $LN$ is layer normalization, $Q$ is the query, $K$ is the key, and $V$ is the value. Afterward, we pair $F_T^{''}$ and $F_S$ together and send them the domain classifier and multi-agent fusion module. 

\subsection{Domain Classifier}
We use the $H$-divergence~\cite{chen2018domain} to measure the divergence between $F_T^{''}$ and $F_S$. Let us denote $X$ as a feature map that may come from the source or target domain and $h: X \rightarrow \{0, 1\}$ a domain classifier, which tries to predict source domain sample $X_S$ as 0 and target domain sample $X_T$ as 1. In our paper, the domain classifier comprises two convolution layers. Suppose $H$ is the hypothesis space for the domain classifier and $G$ is the combination of our learnable resizer and sparse cross-domain transformer, then $G$ needs to be optimized towards the following objective:
\begin{equation}
   \underset{G}{\mathbf{max}}\;\underset{h \in H}{\mathbf{min}}\;(\textbf{E}_S(h(X))+\textbf{E}_T(h(X))
\end{equation}
where $\textbf{E}_S(h(X))$ and $\textbf{E}_T(h(X))$ are the domain classification error over the source domain and target domain respectively and $X$ is produced by $G$. This optimization can be achieved in an adversarial training manner by a gradient reverse
layer~(GRL)~\cite{ganin2015unsupervised}.

\subsection{Multi-Agent Fusion}
Our MPDA framework is very flexible and can integrate most of the multi-agent  fusion algorithms. In this work, we select a state-of-the-art model, V2X-ViT~\cite{xu2022v2xvit}, as our multi-agent fusion algorithm. V2X-ViT employs a heterogeneous multi-agent self-attention block and a multi-scale windowed attention block sequentially to intelligently fuse the different agents' features. To achieve the best performance, besides learning to fool the domain classifier, $G$ also targets to directly optimize the detection performance. Let us denote $M$ as the multi-agent fusion algorithm, then the second training objective for $G$ is:

\begin{equation}
   \underset{G, M}{\mathbf{min}}\;(\textbf{E}_D(V)), \quad V=M(F_S, F_T^{''}), 
\end{equation}
where $\textbf{E}_D(V)$ is the 3D detection error and $V$ is the fused feature with shape of ${1 \times H_S \times W_S \times C_S}$ . 

\subsection{Loss}
For 3D object detection, we use the smooth L1 loss for bounding box regression and focal loss~\cite{lin2017focal} for classification. For the domain classifier, we utilize cross-entropy loss to learn domain-invariant features. The final loss is the combination of detection and domain adaptation loss:
\begin{equation}
    L = \alpha L_{det} + \beta L_{domain},
\end{equation}
where $\alpha$ and $\beta$ are the balance coefficients within range $[0, 1]$.

\section{Experiments}
\subsection{Dataset}
We conduct experiments on the public large-scale V2X perception dataset V2XSet~\cite{xu2022v2xvit}. V2XSet is collected together by the high-fidelity simulator CARLA~\cite{Dosovitskiy17} and cooperative driving automation frame~\cite{xu2021opencda}. It provides LiDAR data from different autonomous vehicles and roadside intelligent infrastructure at the same timestamp and scenario. In total, V2XSet has 11,447 frames and can be split into  6,694/1,920/2,833 frames for 
training/validation/testing respectively. 

\subsection{Experiments Setup}
\noindent\textbf{Evaluation metrics.} We evaluate the performance of our proposed framework by the final 3D detection accuracy. Similar to previous works in this area~\cite{xu2022opv2v, xu2022v2xvit}, we set the evaluation range as $x \in [-140, 140]$ meters, $y \in [-40, 40]$ meters and measure the accuracy  with Average Precisions (AP) at Intersection-over-Union (IoU) threshold of 0.7. 

\noindent\textbf{Evaluation protocols}
During training, we randomly select one agent as the ego agent. During testing, we choose a fixed one as the ego for each scenario. We estimate our model under three distinct settings: 
\begin{enumerate}
    \item \textit{Normal scenario:} In this scenario, all ego agents and other agents use PointPillar~\cite{lang2019pointpillars} with identical parameter as the detection backbone, named $p_0$. \textbf{Among all experiments, the ego vehicle will always have $p_0$ as the backbone.}
    
    \item \textit{Hetero scenario 1:} During training, ego agents use PointPillar $p_0$, whereas other agents employ $p_1$, which also belongs to the PointPillar family but with heterogeneous configurations including voxelization resolutions and the number of convolution layers. To assess the generalization probability of the proposed MPDA, another trained PointPillar model $p_2$ will be used for testing, which has different parameters from any training model.
    
    \item \textit{Hetero scenario 2:} We assume even the model types are heterogeneous in this scenario. All ego vehicles are still trained based on $p_0$, and other agents are based on the different detection model  SECOND~\cite{yan2018second} $s_0$. During the testing stage, we use another trained SECOND model $s_1$ with distinct parameters with $s_0$.

    % \item \textit{Hetero Setting 3:} In this setting, we aim to assess whether our framework can still work for the model architecture that it never sees. During training, the collaborators randomly choose $p_0$, $p_1$, $p_2$, $s_0$, or $s_1$ as backbones, and during testing, they will use an unseen backbone VoxelNet~\cite{zhou2018voxelnet} $v_0$.
    % \textcolor{red}{\textit{Mixed-model scenario}: In this scenario, all ego agents are trained based on PointPillar $p_0$, while other agents are trained by randomly choosing $p_0$, $p_1$, $p_2$, $s_0$, or $s_1$ as backbones. To test our framework for unseen model, the VoxelNet~\cite{zhou2018voxelnet} $v_0$ is used to testing our proposed method.}
\end{enumerate}
To ensure all the backbones are trained properly, we first assume that all agents are equipped with the same backbone and combine it with the V2X-ViT model~\cite{xu2022v2xvit} to perform 3D detection. As shown in 
Table~\ref{tab:hom}, all backbones have achieved reasonable accuracy. We also demonstrate partial parameters of various backbones in Tabel~\ref{tab:param}, and there are noticeable differences between them in terms of voxel resolution, LiDAR cropping range, and the number of convolutional layers.

\begin{table}[h]
    \centering
    \def\xwidth{0.5}
    \caption{\textbf{Detection backbone models' performance on the testing set without domain gap.} We assume all agents have the same detection model in this experiment.}
    \begin{tabular}{c|c|c|c|c|c}
    % \toprule
         \cellcolor{lightgray}{} &\cellcolor{lightgray}{$p_0$} &\cellcolor{lightgray}{$p_1$}  &\cellcolor{lightgray}{$p_2$} &\cellcolor{lightgray}{$s_0$} &\cellcolor{lightgray}{$s_1$}   \\
        \toprule
         AP@0.7 &  71.2 & 68.3 & 70.1 & 74.5 & 77.0  \\
         \bottomrule
    \end{tabular}
    \label{tab:hom}
\end{table}

\begin{table}[h]
    \centering
    \caption{\textbf{Parameters of different detection backbone.}}
    \begin{tabular}{c|c|c|c|c|c}
    % \toprule
         \cellcolor{lightgray}{Backbone} &\cellcolor{lightgray}{{\begin{tabular}[c]{@{}c@{}}Voxel\\ Resolution \end{tabular}} } &\cellcolor{lightgray}{\begin{tabular}[c]{@{}c@{}}Half Lidar\\ Cropping Range~(x\&y)\end{tabular}}   &\cellcolor{lightgray}{\begin{tabular}[c]{@{}c@{}}\# of 2D\&3D\\ CNN Layers\end{tabular}}    \\
        \toprule
         $p_0$ & 0.4, 0.4, 4 & 140.8 \& 38.4 & 19 \& 0 \\
         $p_1$ &  0.8, 0.6, 4  & 140.8 \& 38.4 &  16 \& 0 \\
         $p_2$ &  0.6, 0.6, 4 & 153.6 \& 38.4 & 17 \& 0  \\
         $s_0$ &  0.2, 0.2, 0.2 & 140.8 \& 41.6 &  12 \& 12
         \\
         $s_1$ &  0.1, 0.1, 0.1 & 140.8 \& 41.6 &  13 \& 13 \\ 
         \bottomrule
    \end{tabular}
    \label{tab:param}
\end{table}

\noindent\textbf{Compared methods:} We consider \textit{No Fusion} as the baseline, which does not involve any collaboration in the system. To demonstrate the significant effect of the domain gap, we first directly use the pre-trained model provided by \cite{xu2022v2xvit} and simply apply bilinear interpolation with the channel dropping technique to align the dimensions. We then let the pre-trained model finetune on \textit{Hetero1} and \textit{Hetero2} scenarios to make the comparison fair since our framework will see features from different domains in training.  To show the effectiveness of the two critical components in our framework, we first only add the learnable resizer. Then we add the sparse cross-domain transformer as well to be our complete framework, MPDA.  We will also compare with \textit{Late Fusion} method, which directly transmits the detected 3d bounding box along with the confidence score and merges all the overlapped predictions according to the sorted confidence scores. Though \textit{Late Fusion} does not have the domain gap issue like intermediate fusion, it still suffers from the confidence score discrepancy issue, e.g., different models can have diverse confidence estimation biases.  

\noindent\textbf{Implementation details:} For the multi-agent fusion method, we follow the same hyperparameters for V2X-ViT as its original implementation in \cite{xu2022v2xvit}. For all backbones training, we use Adam~\cite{loshchilov2017decoupled}
as the optimizer, decay the learning rate by 0.1 for every 10 epochs with an initial learning rate of 0.001. The coefficient of detection loss $L_{det}$ is set to $1.0$ and that of domain classification loss $L_{domain}$ is set to $0.1$.

\begin{table}[h]
    \centering
    \def\xwidth{0.3}
    \caption{\textbf{3D detection performance in Normal scenario (w/o domain gap) and Hetero scenarios (w/ domain gap).} We show the Average Precision~(AP) at IoU=0.7. DC stands for domain classifier. * notes that we do not use the domain classifier when training on the normal scenario.}
    \begin{tabular}{c|c|c|c}
    % \toprule
         \cellcolor{lightgray}{Method} &\cellcolor{lightgray}{Normal} &\cellcolor{lightgray}{Hetero 1}  &\cellcolor{lightgray}{Hetero 2}   \\
        \toprule
         No Fusion &  40.2 & 40.2 & 40.2  \\
         Late Fusion & 60.2 & 51.7 & 52.8 \\
         V2X-ViT & 71.2 & \underline{26.7} & \underline{34.5} \\
         V2X-ViT~(finetuned) & 71.2 & 48.6 & 64.8 \\
                  \midrule
         V2X-ViT + Resizer & 72.3  & 54.8 & 72.1 \\
        V2X-ViT + MPDA (w/o DC) & 73.4 & 56.3 &  72.5\\
         V2X-ViT + MPDA & \textbf{73.4*} & \textbf{57.6} & \textbf{73.3}\\

         \bottomrule
    \end{tabular}
    \label{tab:main}
\end{table}

\begin{figure}
    \centering
    \includegraphics[width=0.7\linewidth]{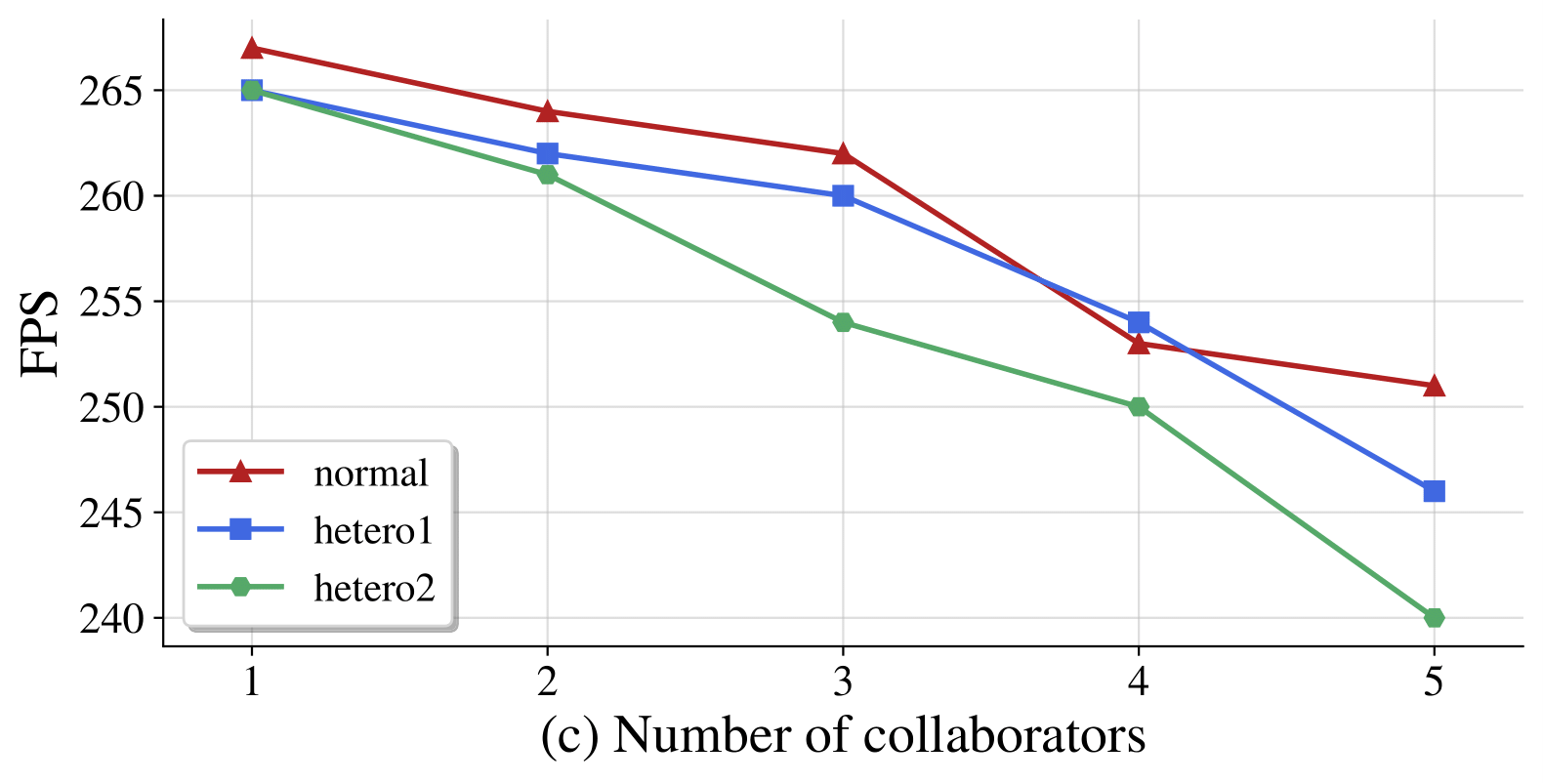}
    \caption{\textbf{Inference speed of MPDA under different settings.}}
    \label{fig:inference}
\end{figure}

\subsection{Quantitaive Evaluation}
\noindent\textbf{Major performance analysis:} Table~\ref{tab:main} depicts the performance comparison of various methods on \textit{Normal},  \textit{Hetro1}, and \textit{Hetero2} settings, respectively. Under the \textit{Normal} scenario, all methods exceed the baseline \textit{No Fusion} by a large margin. Nevertheless, the results are different when the models deployed on the agents are heterogeneous. The pre-trained V2X-ViT drops to \underline{26.7\%} and \underline{34.5\%} on \textit{Hetero1} and \textit{Hetero2} respectively, which is even much lower than the single agent perception system. \textbf{This dramatic performance drop indicates highly negative impacts by the domain gap.} After directly finetune on \textit{Hetero1} and \textit{Hetero2}, V2X-ViT's performance increases though still not satisfying and lower than \textit{Late Fusion} in \textit{Hetero1}. On the contrary, our MPDA has achieved 57.6\%, and 73.3\% on the two heterogeneous settings, which performs favorably against other methods and significantly outperforms \textit{No Fusion}'s baseline. Note that the performance on \textit{Hetero1} is relatively lower for all methods. A potential reason for it is $p_2$'s voxel resolution, and the LiDAR cropping range is quite distinct from $p_0$ and $p_1$, which makes the adaption challenging. With the deployment of our framework, the accuracy of V2X-ViT increases by 9\% and 8.5\% on \textit{Hetero1} and \textit{Hetero2} respectively. We also found that our MPDA can enhance the performance on \textit{Normal} setting as well by 2.2\%, which attributes the capability of our resizer and sparse cross-domain transformer to help generate more robust feature representations.

\begin{figure*}[!t]
\centering
    \begin{subfigure}[c]{0.29\linewidth}
        \centering{\includegraphics[width=1\linewidth]{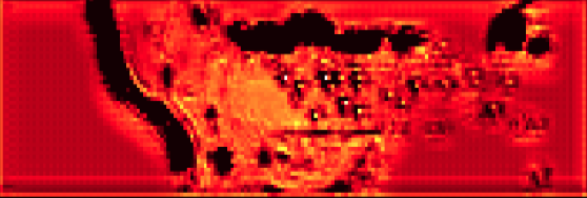}}
        \centering{\includegraphics[width=1\linewidth]{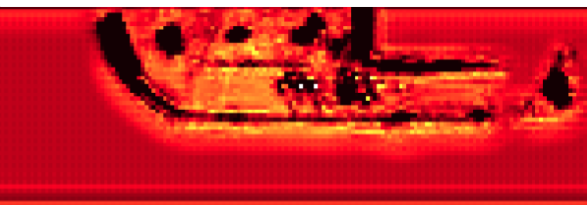}}

        \caption{ego's feature}
        \label{fig:output-a}
    \end{subfigure}
    \begin{subfigure}[c]{0.29\linewidth}
        \centering{\includegraphics[width=1\linewidth]{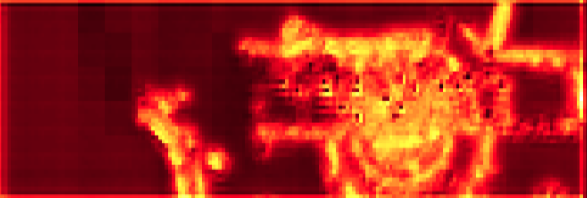}}
        \centering{\includegraphics[width=1\linewidth]{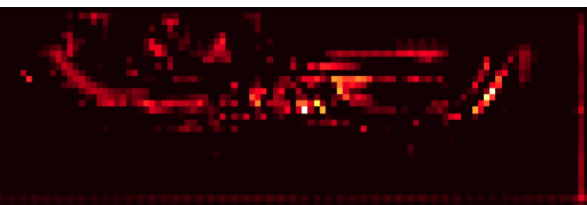}}
        \caption{collaborator's feature before MPDA}
        \label{fig:output-b}
    \end{subfigure}
    \vspace{1mm}
    \begin{subfigure}[c]{0.29\linewidth}
        \centering{\includegraphics[width=1\linewidth]{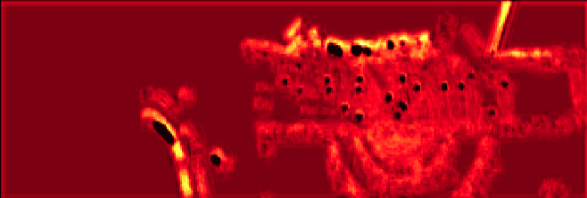}}
        \centering{\includegraphics[width=1\linewidth]{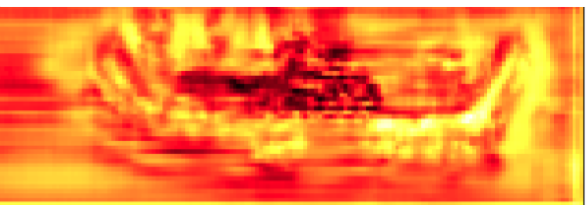}}
        \caption{collaborator's feature after MPDA}
        \label{fig:output-c}
    \end{subfigure}

    \caption{\textbf{Visalzation of intermediate features before and after domain adaption.} From left to right: (a) ego's feature, (b) collaborator's feature before domain adaption, (c) collaborator's feature after domain adaption. Row 1 is the \textit{Hetero1} scenario where ego and others both use PointPillar, but the parameters differ. Row 2 is the \textit{Hetero2} scenario where ego uses PointPillar, and others use SECOND. It is obvious that after domain adaption, others' intermediate features have more similar patterns as ego's.}
    \label{fig:visda}
    \vspace{-2mm}
\end{figure*}

\begin{figure*}[thbp]%
  \centering%
  \subfloat[Original V2X-ViT in Hetero1]{%
    \resizebox{0.28\linewidth}{!}{
      \begin{tikzpicture}%
        \node at(0.0,0.0){\fbox{\includegraphics[width=\linewidth]{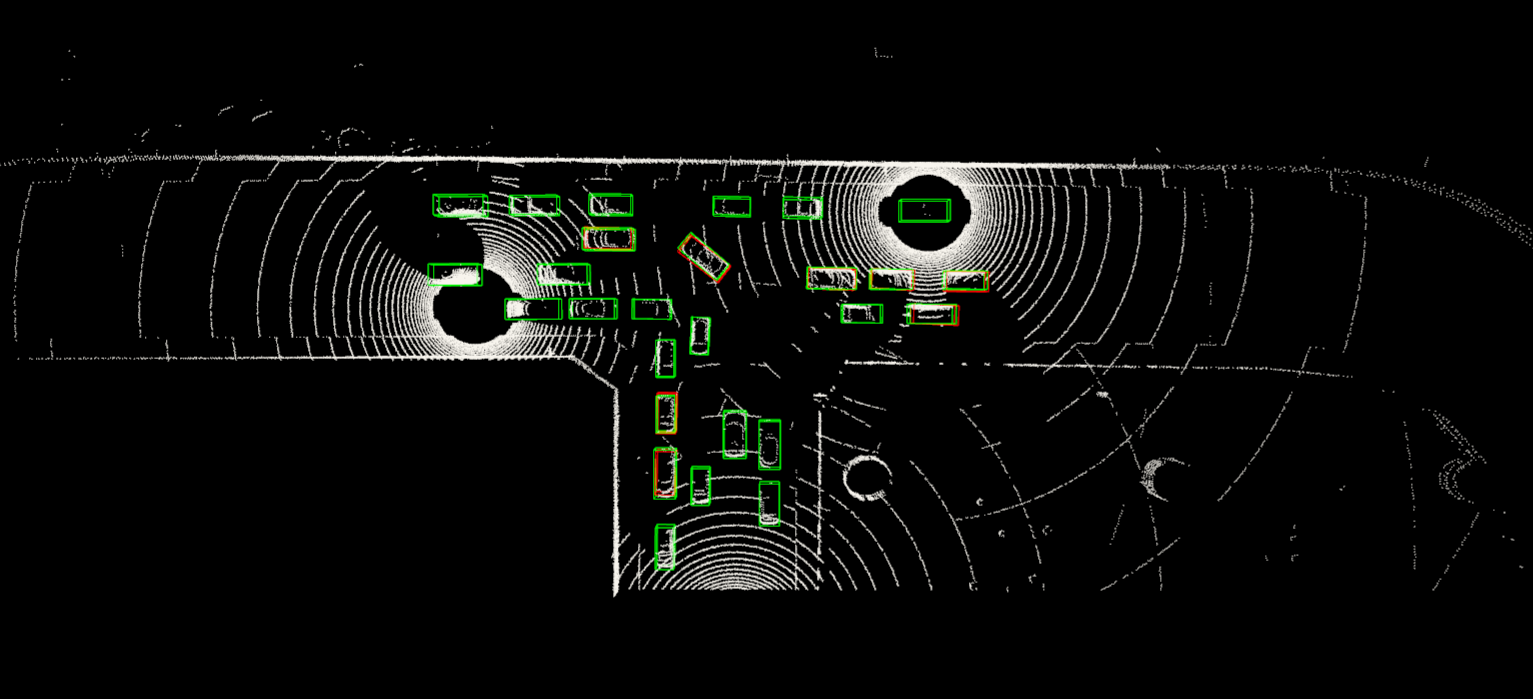}}};%
      \end{tikzpicture}%
    }%
  }%
  \subfloat[Finetuned V2X-ViT in Hetero1]{%
    \resizebox{0.28\linewidth}{!}{
      \begin{tikzpicture}%
        \node at(0.0,0.0){\fbox{\includegraphics[width=\linewidth]{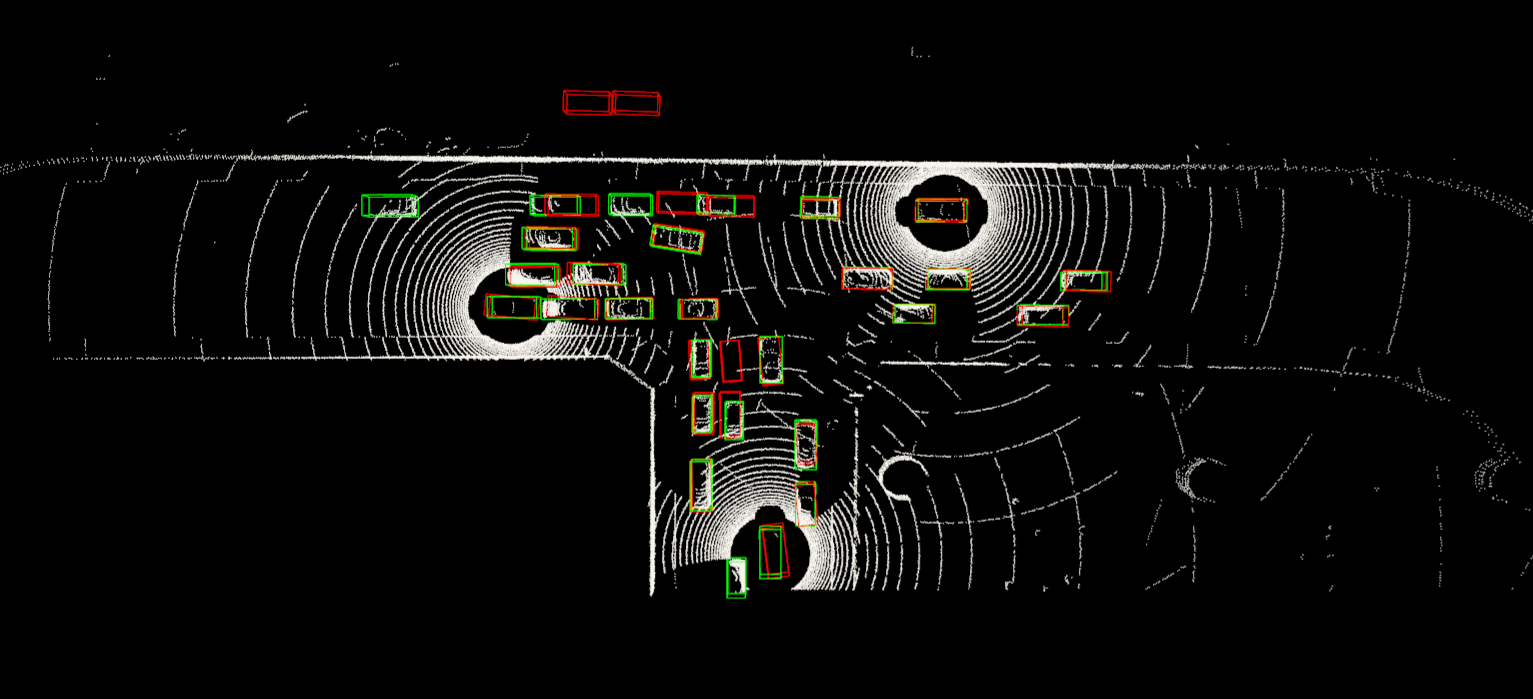}}};%
      \end{tikzpicture}%
    }%
  }%
  \subfloat[V2X-ViT + MPDA in Hetero1]{%
    \resizebox{0.28\linewidth}{!}{
      \begin{tikzpicture}%
        \node at(0.0,0.0){\fbox{\includegraphics[width=\linewidth]{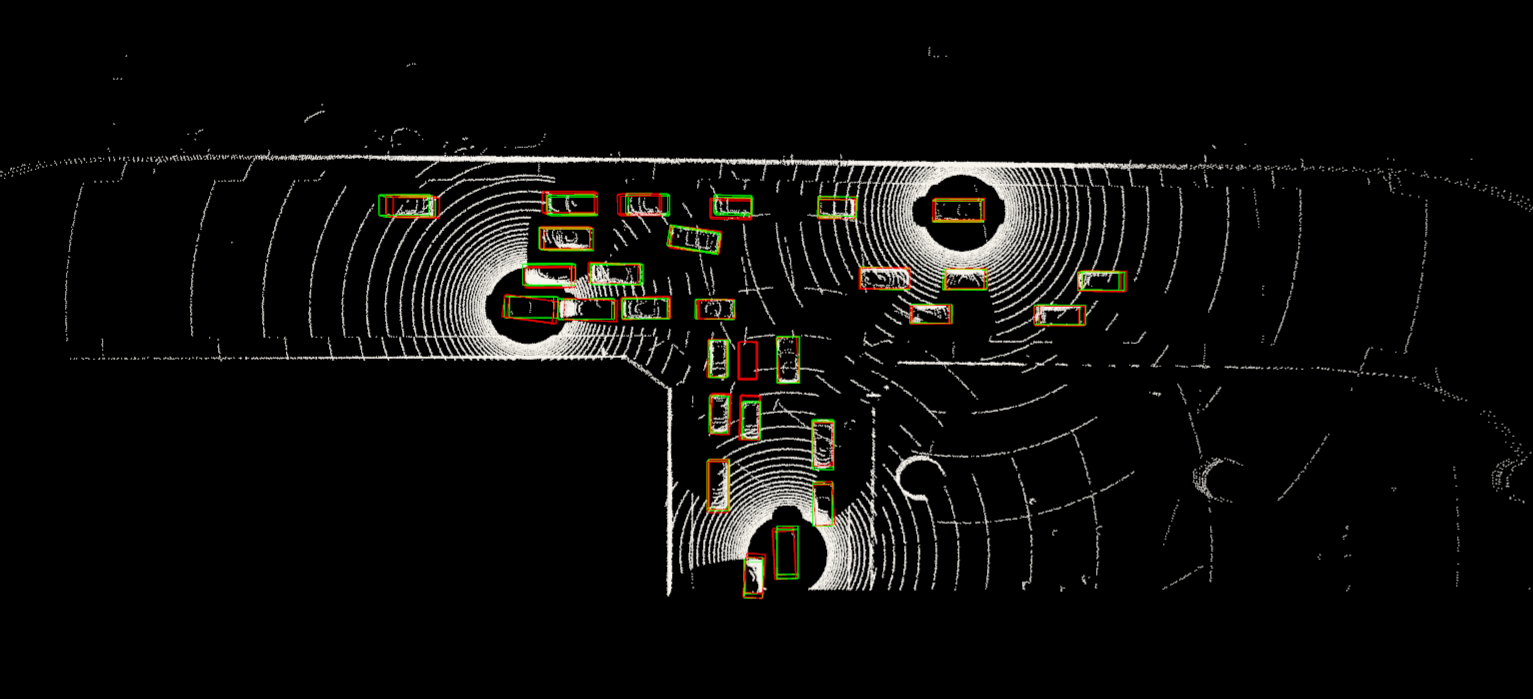}}};%
      \end{tikzpicture}%
    }%
  }\\%
 
  \subfloat[Original V2X-ViT in Hetero2]{%
    \resizebox{0.28\linewidth}{!}{
      \begin{tikzpicture}%
        \node at(0.0,0.0){\fbox{\includegraphics[width=\linewidth]{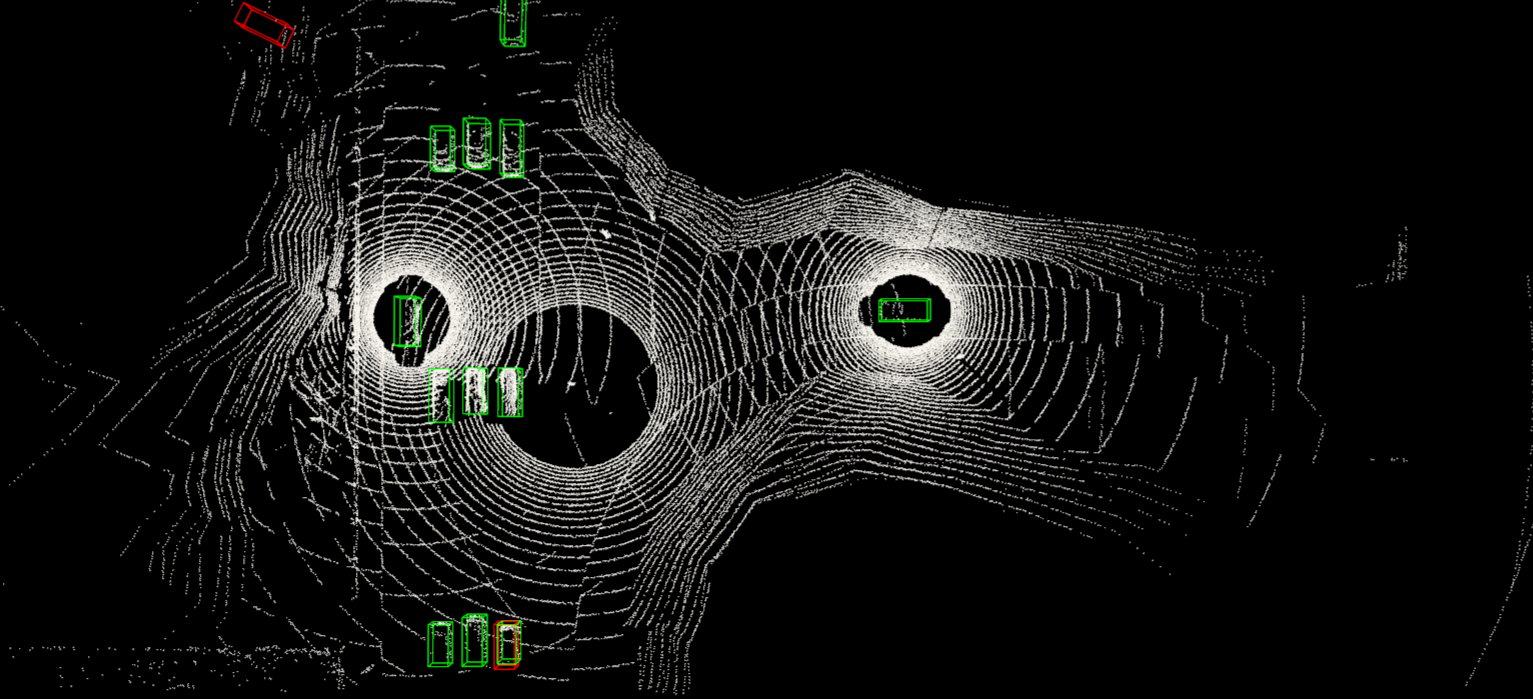}}};%
      \end{tikzpicture}%
    }%
  }%
  \subfloat[Finetuned V2X-ViT in Hetero2]{%
    \resizebox{0.28\linewidth}{!}{
      \begin{tikzpicture}%
        \node at(0.0,0.0){\fbox{\includegraphics[width=\linewidth]{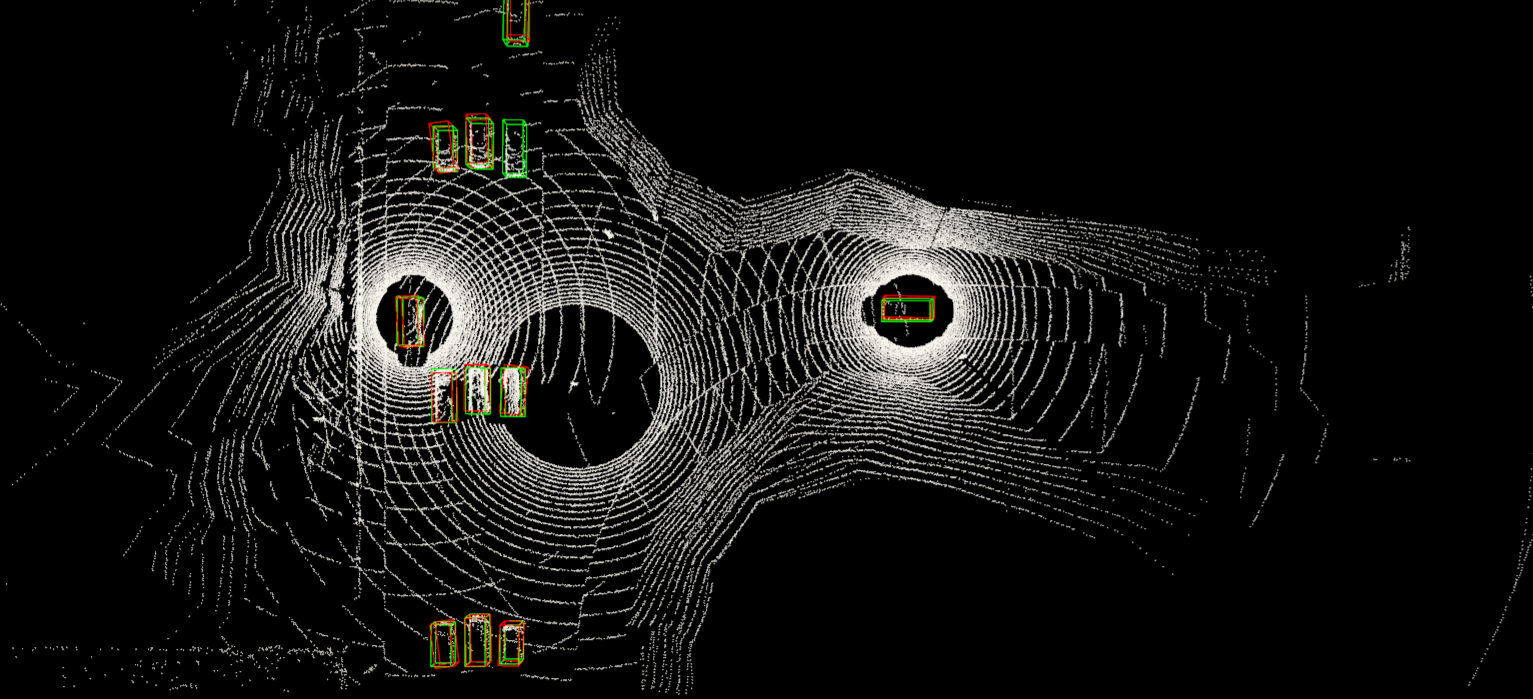}}};%
      \end{tikzpicture}%
    }%
  }%
  \subfloat[V2X-ViT + MPDA in Hetero2]{%
    \resizebox{0.28\linewidth}{!}{
      \begin{tikzpicture}%
        \node at(0.0,0.0){\fbox{\includegraphics[width=\linewidth]{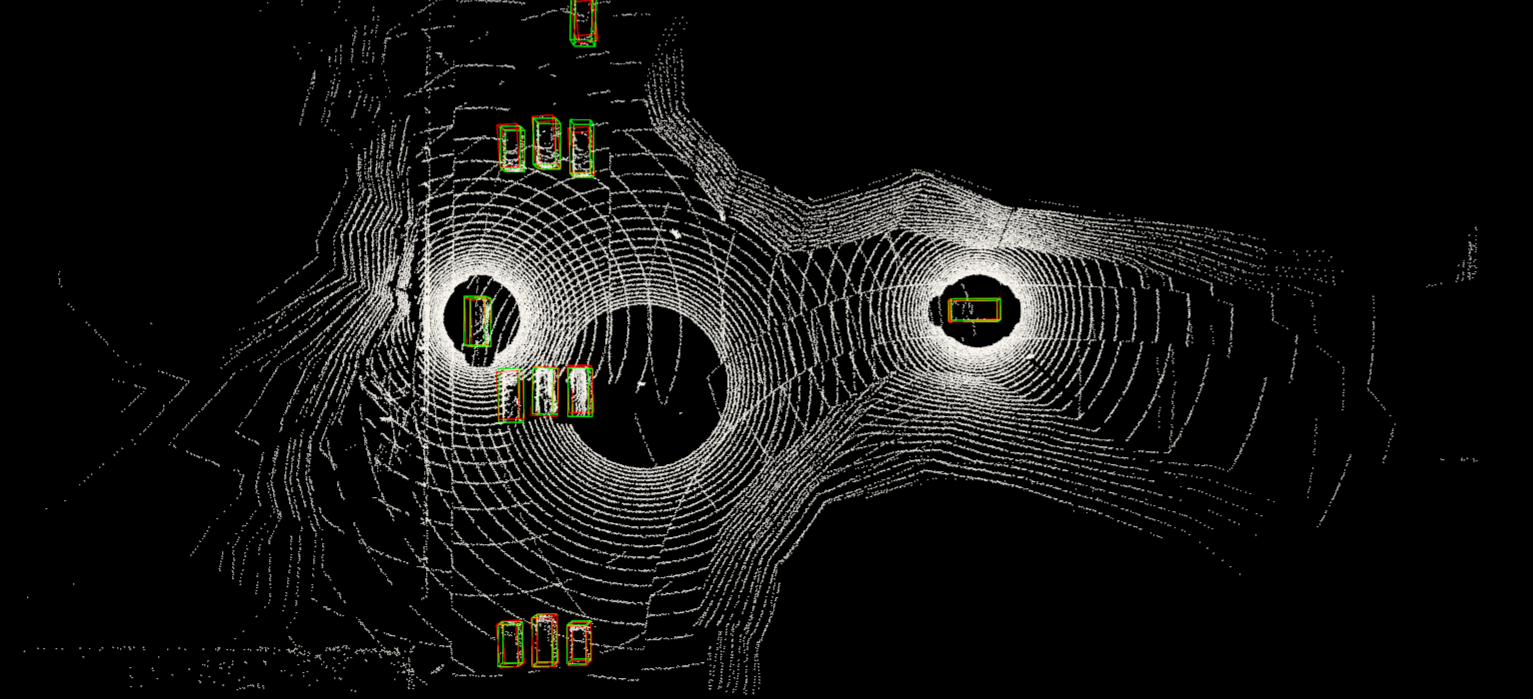}}};%
      \end{tikzpicture}%
    }%
  }%
  \caption{\textbf{3D detection visualization.} \textcolor{green}{Green} and \textcolor{red}{red} 3D bounding boxes represent the \textcolor{green}{ground truth} and \textcolor{red}{prediction} respectively. With our MPDA, the detection results are clearly more accurate.}%
  \label{fig:detect}%
\end{figure*}

\noindent\textbf{Main component analysis:} As 
Table~\ref{tab:main} describes, all of our designed components in MPDA have contributed to  more accurate detections. Adding the learnable resizer improves the detection performance by 6.2\% and 7.3\% under two heterogeneous settings. The sparse cross-domain transformer combined with the domain classifier can further increase the AP by 2.8\% and 1.2\%.

\noindent\textbf{Inference time:} Real-time performance is critical for real-world deployment. Thus, here we calculate the inference speed of our proposed MPDA framework under different scenarios concerning various collaborator numbers. As Fig.~\ref{fig:inference} shows, our MPDA can always achieve more than 200FPS under different settings, indicating our design's efficiency.

\subsection{Qualitative Evaluation}
\noindent \textbf{Domain adaption visualization:} Similar to \cite{zagoruyko2016paying}, we sum up all the absolute values of all channels to visualize the feature maps to investigate their patterns. As Fig.~\ref{fig:visda} shows, without any domain adaption, there are noticeable gaps between the ego agent's and other collaborators' features. After applying our MPDA, the converted features become more similar to the ego's, which visually proves the effectiveness of MPDA.

\noindent \textbf{3D detection visualization:} We visually compare different methods in the same scenario under two heterogeneous settings and show the result in Fig.~\ref{fig:detect}. Obviously, without seeing any features from different backbones, the pre-trained V2X-ViT model provided by the authors from \cite{xu2022v2xvit} has many missing detections. After directly finetuning under \textit{Hetero1} and \textit{Hetero2} settings, the results get improved, but there still exist noticeable missing detections, false positives, and large displacement. On the contrary, our MPDA has a more robust performance, detecting most of the objects and predicting accurate bounding box positions.

\section{CONCLUSIONS}
This paper is the first work that investigates the domain gap issue in multi-agent perception. Based on the analysis, we propose the first multi-agent perception domain adaption framework, which mainly contains a learnable feature resizer and a sparse cross-domain transformer. Extensive experiments on the V2XSet dataset prove that our framework can effectively bridge the domain gap. In the future, we will combine robust generative representation learning techniques such as Diffusion~\cite{Yang2022DiffusionMA} and conduct real-world field experiments on this practical issue. 
%0.548

\section{Acknowledgement}
This work is part of the OpenCDA Ecosystem~\cite{10045043} and is supported by the  Federal Highway Administration with Grant number 693JJ321C000016.

% \addtolength{\textheight}{-12cm} 
\addtolength{\textheight}{-3cm}   % This command serves to balance the column lengths
                                  % on the last page of the document manually. It shortens
                                  % the textheight of the last page by a suitable amount.
                                  % This command does not take effect until the next page
                                  % so it should come on the page before the last. Make
                                  % sure that you do not shorten the textheight too much.

%%%%%%%%%%%%%%%%%%%%%%%%%%%%%%%%%%%%%%%%%%%%%%%%%%%%%%%%%%%%%%%%%%%%%%%%%%%%%%%%

%%%%%%%%%%%%%%%%%%%%%%%%%%%%%%%%%%%%%%%%%%%%%%%%%%%%%%%%%%%%%%%%%%%%%%%%%%%%%%%%

%%%%%%%%%%%%%%%%%%%%%%%%%%%%%%%%%%%%%%%%%%%%%%%%%%%%%%%%%%%%%%%%%%%%%%%%%%%%%%%%

\bibliographystyle{IEEEtran}
\bibliography{IEEEfull}

\end{document}